\documentclass[letterpaper, 10 pt, conference]{URL-ieeeconf}
\IEEEoverridecommandlockouts    
\usepackage{graphics}           
\usepackage{times}              
\usepackage{amsmath}            
\usepackage{amssymb}            
\usepackage{graphicx}
\usepackage{algorithm}
\usepackage[noend]{algpseudocode}
\usepackage{booktabs}
\usepackage{color}
\usepackage{multirow}
\usepackage{subcaption}
\usepackage{rotating}
\usepackage{cite}
\definecolor{instructioncolor}{rgb}{.5,.5,.5}

\usepackage[font=small]{caption}

\def\eqref#1{(\ref{#1})}

\def\vsfig{\vspace{-0.3cm}}
\def\vstab{\vspace{-0.3cm}}

\newcommand{\rom}[1]{\uppercase\expandafter{\romannumeral #1\relax}}

\makeatletter
\usepackage{xspace}
\DeclareRobustCommand\onedot{\futurelet\@let@token\@onedot}
\def\@onedot{\ifx\@let@token.\else.\null\fi\xspace}

\def\etal{{\textit{et al}}\onedot}
\makeatother

\usepackage{array}
\newcolumntype{L}[1]{>{\raggedright\let\newline\\\arraybackslash\hspace{0pt}}m{#1}}
\newcolumntype{C}[1]{>{\centering\let\newline\\\arraybackslash\hspace{0pt}}m{#1}}
\newcolumntype{R}[1]{>{\raggedleft\let\newline\\\arraybackslash\hspace{0pt}}m{#1}}

\title{\LARGE \bf Robust Structureless Monocular Visual-Inertial Initialization\\Exploiting Line Features and Vanishing Points}

\author{Junwan Choi$^1$, Woongrae Jo$^1$, Dong-Uk Seo$^1$, Jinwoo Jeon$^1$, \textit{Student Member, IEEE},\\
and Hyun Myung$^{1*}$, \textit{Senior Member, IEEE}
\vspace{0.1cm}
  \thanks{$^*$Corresponding author: Hyun Myung}
  \thanks{$^{1}$Junwan Choi, Woongrae Jo, Dong-Uk Seo, Jinwoo Jeon, and Hyun Myung are with the School of Electrical Engineering, KAIST (Korea Advanced Institute of Science and Technology), Daejeon, 34141, Republic of Korea. {\tt\scriptsize \{joseph3155, woongrae8879, dongukseo, jinuok, hmyung\}@kaist.ac.kr} \hfill \break
  }
}

\makeatletter
\let\NAT@parse\undefined
\makeatother
\usepackage{hyperref}
\usepackage{cleveref}

\usepackage{multirow}
\usepackage{booktabs}
\usepackage[table]{xcolor}
\usepackage{colortbl}
\usepackage{array}
\usepackage{gensymb}

\begin{document}
\maketitle
\thispagestyle{empty}
\pagestyle{empty}

  

\begin{abstract}

Accurate initialization is essential for reliable visual-inertial odometry (VIO), but it is often ill-conditioned under degenerate motions.
Existing methods typically require restrictive excitation motions to ensure sufficient observability or rely on computationally expensive 3D structure reconstruction, limiting efficient and practical deployment.
To address these limitations, we propose \textit{SLIM-init}, a structureless monocular VIO initializer that directly exploits geometric constraints from tracked 2D line features without explicit 3D landmark reconstruction.
Specifically, SLIM-init leverages line-derived vanishing points (VPs) as translation-invariant orientation cues to provide robust rotation-only constraints under degenerate scenarios such as low-parallax or translation-dominant motions. It further incorporates a line epipolar residual to constrain translation and a line-normal projection residual to improve the conditioning of linear alignment, enhancing the accuracy and robustness of initial state estimation.
Extensive experiments on a public benchmark and challenging custom degenerate-motion sequences demonstrate improved accuracy and robustness over state-of-the-art initialization methods. The source code is available at: \url{https://github.com/cjunwan/SLIM-init}.

\end{abstract}

\section{Introduction}
\label{sec:intro}








%
%
%
%

Visual-inertial odometry (VIO) and visual simultaneous localization and mapping (SLAM) are fundamental for state estimation in GPS-denied environments, enabling applications in augmented/virtual reality, drones, and robotics. By fusing measurements from a camera and an inertial measurement unit (IMU), monocular VIO provides a low-cost and lightweight solution for accurate 6-DoF motion tracking. Most monocular VIO systems~\cite{qin2018vins, campos2021orb, leutenegger2015keyframe, geneva2020openvins} start with an explicit initialization stage to estimate key parameters such as metric scale, gravity direction, initial velocity, and IMU biases, which critically affect the stability and accuracy of subsequent estimation~\cite{hartley2003multiple, kelly2011visual, lynen2015get}.


In practical deployments, reliable initialization is essential not only at startup but also for on-the-fly recovery from tracking failures or estimator divergence~\cite{lynen2015get, qin2017robust, scheiber2021midair, song2024dynavinspp}. Achieving sufficient observability in both cases, however, often requires operational constraints, such as specific excitation motions~\cite{kelly2011visual, qin2018vins, geneva2020openvins} or prolonged bootstrapping~\cite{mur2017visual, huang2018online}. Although effective in controlled settings, these requirements can hinder usability and limit deployment under unconstrained conditions, motivating robust and efficient initialization.


Nevertheless, robust initialization under degenerate motions remains challenging. Existing approaches often address limited observability using richer geometric constraints from multi-view reconstruction and nonlinear optimization over a large number of landmarks, which can be costly for real-time deployment~\cite{leutenegger2015keyframe, qin2017robust, mur2017visual}. Moreover, approaches relying primarily on point-parallax constraints can become ill-conditioned under low-parallax or translation-dominant motions, precisely where robust initialization is most needed.



To complement point-parallax constraints, several works incorporate structural cues from line features, which provide useful geometric information in man-made or low-texture environments~\cite{liu2022integrating, xie2024pointline}. However, existing line-augmented designs still rely on explicit 3D structure estimation (e.g., maintaining 3D line landmarks or estimating line depths), inheriting reconstruction fragility under limited parallax while adding computational overhead. Recent structureless initializers~\cite{he2023rotation, song2025improving} reduce reconstruction costs through low-dimensional optimization and efficient linear alignment, but their predominantly point-based formulations remain susceptible to ill-conditioning under limited observability, leaving robustness an unresolved challenge.


To address these limitations, we propose \textit{SLIM-init}, a robust and computationally efficient monocular VIO initializer that incorporates additional geometric constraints without reintroducing explicit 3D reconstruction. Our key insight is that line-derived cues can improve initialization in a strictly structureless manner. We leverage line-derived vanishing points (VPs) as translation-invariant orientation cues for rotation-only gyroscope bias estimation and augment translation-based linear alignment with a line epipolar residual and a line-normal projection residual to improve scale, gravity, and velocity estimation under degenerate motions. Together, these components improve accuracy and robustness with marginal overhead, enabling real-time initialization. Our main contributions are summarized as follows:

\begin{itemize}
 \item We propose a strictly structureless monocular VIO initialization method that incorporates line-derived structural constraints directly from 2D line features, without any 3D reconstruction or line depth estimation. To the best of our knowledge, this is the first structureless line integration in monocular VIO initialization.

 \pagebreak
 
 \item We incorporate a translation-invariant VP consistency residual to strengthen gyroscope bias estimation under degenerate motions. We also introduce a line epipolar residual and a line-normal projection residual to enhance the linear alignment for scale, gravity, and velocity.

 \item Experiments on a public benchmark     and challenging custom degenerate-motion sequences show improved accuracy and robustness over state-of-the-art initializers while running in real time. The source code is publicly released to facilitate future research.
\end{itemize}

\section{Related Work}
\label{sec:related}




\subsection{Structure-based Visual-Inertial Initialization}
Classical visual-inertial initialization methods are predominantly structure-based, relying on explicit 3D landmark reconstruction for initial state estimation. Based on how visual and inertial information are combined, these approaches can be broadly categorized into loosely-coupled and tightly-coupled paradigms.

Loosely-coupled methods~\cite{yang2017monocular, qin2017robust, qin2018vins} first perform visual structure-from-motion (SfM) to recover camera poses and a 3D point cloud up to an unknown scale, and subsequently align the visual reconstruction with IMU preintegration to estimate scale and gravity. While effective when the visual reconstruction is stable, these pipelines degrade under degenerate scenarios such as low-texture scenes or low parallax, where point triangulation becomes poorly conditioned.

Tightly-coupled approaches~\cite{bloesch2015robust, shen2015tightly, leutenegger2015keyframe} integrate visual and inertial measurements through filtering or joint optimization, reducing reliance on a separate SfM stage. However, their performance can degrade with noisy IMU data due to limited visual refinement of bias estimates~\cite{martinelli2011vision, kaiser2016simultaneous}. Recent efforts~\cite{campos2020inertial, evangelidis2021revisiting, zuniga2021analytical} have been made to improve robustness against sensor noise through uncertainty modeling and compact analytical formulations, but jointly estimating 3D landmarks and inertial quantities remains high-dimensional, computationally costly, and numerically fragile under degenerate conditions~\cite{he2023rotation}.

\subsection{Line-Augmented VIO and Initialization}
To alleviate the limited geometric constraints of point-driven pipelines, line features have been explored as complementary structural cues in VIO and visual SLAM~\cite{pumarola2017plslam, he2018plvio, zou2019structvio, lim2022uvslam, seo2023semidense}. By providing constraints beyond point correspondences, line-derived cues can improve robustness in structured or low-texture scenes where point tracking or triangulation is unreliable.

Most line-based systems rely on explicit structure, including 3D line-landmark optimization~\cite{he2018plvio,zou2019structvio}, line-depth estimation~\cite{liu2022integrating,xie2024pointline}, and semi-dense 3D maps for robust line tracking~\cite{seo2023semidense}. While these methods demonstrate improved accuracy and robustness, explicit structural estimation incurs substantial overhead, hindering efficient real-time initialization. Moreover, many of these pipelines still rely on maintaining a stable visual reconstruction. As an alternative, PLE-SLAM~\cite{he2024pleslam} avoids explicit line reconstruction by augmenting IMU initialization with point/line coplanarity constraints. Nevertheless, it assumes a stereo-provided metric scale, fundamentally differing from the strictly monocular-inertial recovery targeted in this work.

\subsection{Structureless Visual-Inertial Initialization}
To overcome the computational burden and structural dependence of explicit 3D reconstruction, structureless initialization methods have recently gained attention. These methods aim to eliminate 3D landmarks from the state vector and exploit multi-view geometric constraints directly. A representative approach is the rotation-translation decoupled framework~\cite{he2023rotation}, which estimates gyroscope bias using rotation-only constraints~\cite{kneip2013direct} and then recovers velocity, gravity, and scale through global linear alignment~\cite{cai2021pose}. Song~\etal~\cite{song2025improving} further introduced a structureless visual-inertial bundle adjustment to refine the estimates.

Despite their efficiency, existing structureless methods remain primarily reliant on point-feature parallax, leading to weak observability in low-texture or low-parallax regimes. To address this limitation, we incorporate 2D line-derived geometric constraints, such as VP direction consistency, line epipolar geometry, and line-normal projection, to improve initialization accuracy and robustness while preserving real-time performance.

\section{Notation}
\label{sec:notation}

We consider a sliding window of $N$ keyframes indexed by $i\in\{0,\dots,N-1\}$, and define $\mathcal{E}$ as the set of all consecutive keyframe pairs. Frames $w$, $b_i$, and $c_i$ denote the world, IMU/body, and camera frames at the $i$-th keyframe.

A rotation matrix $\mathbf{R}_{ab}\in \mathrm{SO}(3)$ maps coordinates from frame $b$ to frame $a$, and $\mathbf{p}_{ab}^c\in\mathbb{R}^3$ denotes the position of frame $b$ with respect to frame $a$, expressed in the coordinate frame $c$. For each pair $(i,j)\in\mathcal{E}$, IMU preintegration over the time interval $\Delta t_{ij}$ yields the nominal rotation, position, and velocity increments $\Delta\bar{\mathbf{R}}_{ij}$, $\Delta\bar{\mathbf{p}}_{ij}$, and $\Delta\bar{\mathbf{v}}_{ij}$. 

A tilde indicates an up-to-scale quantity, e.g., $\tilde{\mathbf{p}}_{w c_i}^w$ as opposed to its metric counterpart $\mathbf{p}_{w c_i}^w$. The known camera-IMU extrinsics are denoted by $\mathbf{R}_{bc}$ and $\mathbf{t}_{bc}^b$, and $[\cdot]_\times$ denotes the skew-symmetric operator.

\vspace{0.1cm}
\section{SLIM-init: Structureless Line-based Monocular Visual-Inertial Initialization}
\label{sec:method}





\begin{figure*}[t]
    \vspace*{0.05in}
    \centering
    \includegraphics[width=0.97\textwidth]{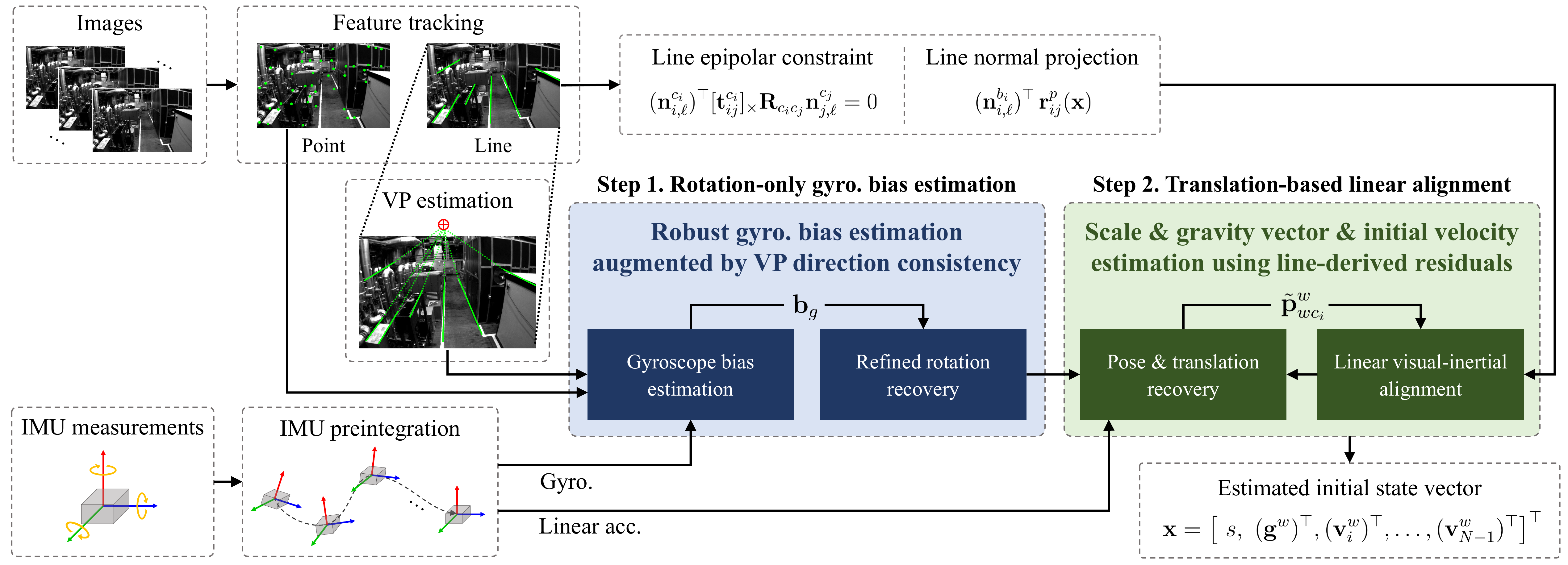}
    \caption{Overall pipeline of SLIM-init. Colored blocks highlight our augmentations: a VP-based consistency residual for robust gyroscope bias estimation (Step~$1$), followed by line epipolar and line-normal projection residuals for translation-based linear alignment (Step~$2$). The entire pipeline is strictly structureless, avoiding depth estimation or any explicit 3D reconstruction.}
    \label{fig:main_pipeline}
    \vspace{-0.3cm}
\end{figure*}

\subsection{Overview}
\label{subsec:overview}

The overall framework of SLIM-init is shown in Fig.~\ref{fig:main_pipeline}. We adopt a rotation-translation decoupled, structureless initialization pipeline based on He~\etal~\cite{he2023rotation}. Given a short window of keyframe images and IMU measurements, the system extracts point features~\cite{shi1994good}, tracks them using a Kanade-Lucas-Tomasi (KLT) optical flow frontend~\cite{lucas1981iterative}, and performs IMU preintegration on the $\mathrm{SO}(3)$ manifold~\cite{forster2016manifold}. Using the processed visual and inertial measurements, initialization then proceeds in two steps: gyroscope bias estimation for relative rotation refinement (Step~$1$), followed by translation recovery and global linear alignment to estimate metric scale, gravity direction, and keyframe velocities (Step~$2$).

To strengthen geometric constraints, we additionally utilize line-derived structural cues. To this end, line features are extracted using EDLines~\cite{akinlar2011edlines} and tracked by optical flow-based prediction with geometric consistency matching, avoiding the use of costly line descriptors. The proposed line-derived structural cues are then integrated into both stages in a strictly structureless manner, bypassing explicit 3D line reconstruction to preserve real-time performance, while addressing the corresponding degeneracies:

\begin{itemize}
    \item \textbf{Step 1 (Rotation-only gyro. bias estimation):} We introduce a translation-invariant VP consistency residual, which provides a rotation-only constraint that stabilizes the bias estimation even when point tracking is degraded.
    \item \textbf{Step 2 (Translation-based linear alignment):} We incorporate a line epipolar residual and a line-normal projection residual. The line epipolar term adds direct translation constraints from line correspondences, while the line-normal projection term provides anisotropic conditioning. Together, they improve the robustness and numerical stability of the linear solver in low-texture or low-parallax scenarios.
\end{itemize}

Detailed formulations of the proposed residuals and their integration are provided in the following subsections.

\subsection{Step 1: Rotation-only Gyroscope Bias Estimation}
\label{subsec:step1}

In the first step, we estimate the gyroscope bias $\mathbf{b}_g$ using rotation-only constraints.
To this end, we introduce a VP direction-based geometric consistency term and minimize the following rotation-only objective:  
\begin{equation}
\begin{aligned}
    \min_{\mathbf{b}_g}\;&
    \sum_{(i,j)\in\mathcal{E}}
    \rho\!\left(
    \sum_{k\in\mathcal{P}_{ij}}
    \big\Vert\mathbf{r}^{\mathrm{pt}}_{ij,k}(\mathbf{b}_g)\big\Vert^2
    \right) \\
    &\quad +
    \lambda_{\mathrm{vp}}
    \sum_{(i,j)\in\mathcal{E}}
    \mathbb{I}^{\mathrm{vp}}_{ij}\,
    \rho\!\left(
    \big\Vert\mathbf{r}^{\mathrm{vp}}_{ij}(\mathbf{b}_g)\big\Vert^2
    \right),
\end{aligned}
\label{eq:bg_objective_new}
\end{equation}
where $\mathbf{r}^{\mathrm{pt}}_{ij,k}$ and $\mathbf{r}^{\mathrm{vp}}_{ij}$ denote the standard point residual~\cite{kneip2013direct} and the proposed VP direction consistency residual, respectively.
Here, $\mathcal{P}_{ij}$ is the set of point features tracked between keyframes $i$ and $j$, and $\rho(\cdot)$ is the Cauchy loss. The adaptive weight $\lambda_{\mathrm{vp}}$ is scaled by the point-to-VP residual count ratio within the sliding window to prevent the joint objective from being dominated by an unequal number of measurements.
Finally, $\mathbb{I}^{\mathrm{vp}}_{ij}$ is a binary indicator that activates the VP term only for reliable VP tracks.

The following subsections detail the estimation and tracking of the dominant VP direction, reliability gating via $\mathbb{I}^{\mathrm{vp}}_{ij}$, and the formulation of the proposed VP consistency residual.

\subsubsection{Dominant VP Direction Estimation and Tracking}
\label{subsubsec:vp_est}

Each tracked 2D line segment $\ell$ at keyframe $i$ is represented by an interpretation-plane normal $\mathbf{n}_{i,\ell}^{c_i}\in\mathbb{S}^2$, while the dominant VP is denoted by a direction vector $\mathbf{d}_i^{c_i}\in\mathbb{S}^2$.
We estimate the dominant VP as the most consistently supported direction across tracked lines, providing a stable structural cue for subsequent reliability gating and the VP consistency residual.

Specifically, the dominant VP direction $\mathbf{d}_i^{c_i}$ is estimated via a vote-and-refine scheme. An initial candidate is obtained by accumulating length-weighted votes on a unit sphere discretized via icosahedral subdivision~\cite{fekete1990rendering}, with voting restricted to the front-facing hemisphere to avoid antipodal redundancy. An inlier set $\mathcal{I}_i$ is formed from lines whose angular error to the candidate is below the threshold $t_{\mathrm{in}}$, and the VP direction is refined by solving a weighted least-squares problem over $\mathcal{I}_i$ as:
\begin{equation}
    \mathbf{d}_i^{c_i}
    =
    \arg\min_{\|\mathbf{d}\|=1}
    \sum_{\ell\in\mathcal{I}_i}
    w_{i,\ell}\big((\mathbf{n}_{i,\ell}^{c_i})^\top\mathbf{d}\big)^2,
\label{eq:vp_fit}
\end{equation}
where $w_{i,\ell}$ is proportional to the line length to emphasize more reliable orientation measurements.

For temporal consistency, we assign each dominant VP a track ID corresponding to the same physical axis.
The ID is propagated by rotating the previous VP direction from frame $i$ to frame $j$ using IMU measurements and matching it with the new observation.
The track age is defined as the number of consecutive frames for which the same ID persists.
To handle intermittent detections or abrupt axis switching, we further employ a reliability gating strategy.

\subsubsection{VP Reliability Gating}
\label{subsubsec:vp_gate}

To mitigate unreliable tracking, a conservative gating mechanism activates VP 
residuals only when they are spatially consistent.
For each consecutive keyframe pair $(i,j)$, a binary indicator $\mathbb{I}^{\mathrm{vp}}_{ij} \in \{0,1\}$ is set to $1$ only if $\mathbf{d}_i^{c_i}$ and $\mathbf{d}_j^{c_j}$ share the same track ID, maintain a minimum track age $A_{\mathrm{min}}$, and satisfy a strict consistency threshold $t_{\mathrm{vp}}$.
Specifically, for each inlier $\ell \in \mathcal{I}_i$, the line-VP consistency error is defined as:
\begin{equation}
    e^{\mathrm{vp}}_{i,\ell} \triangleq (\mathbf{n}_{i,\ell}^{c_i})^\top \mathbf{d}_i^{c_i}, 
\label{eq:vp_inlier_error}
\end{equation}
where $e^{\mathrm{vp}}_{i,\ell}$ represents the geometric misalignment between the line interpretation plane and the estimated VP direction at keyframe~$i$.

\subsubsection{VP Consistency Geometric Residual}
\label{subsubsec:bg_obj}

The standard point-based rotation residual often lacks robustness under low-parallax or low-texture conditions. To augment the rotation-only estimation in such scenarios, we propose the following VP direction consistency residual $\mathbf{r}^{\mathrm{vp}}_{ij}$:
\begin{equation}
    \mathbf{r}^{\mathrm{vp}}_{ij}(\mathbf{b}_g)
    =
    \mathbf{d}_j^{c_j}
    \times
    \big(\mathbf{R}_{c_j c_i}(\mathbf{b}_g)\mathbf{d}_i^{c_i}\big)
    , 
\label{eq:vp_residual_impl}
\end{equation}
where $\mathbf{R}_{c_j c_i}(\mathbf{b}_g)$ denotes the bias-corrected relative camera rotation, given by:
\begin{equation}
    \mathbf{R}_{c_j c_i}(\mathbf{b}_g)
    =
    \mathbf{R}_{b_j c_j}^\top\,
    \mathbf{R}_{b_j b_i}(\mathbf{b}_g)\,
    \mathbf{R}_{b_i c_i}. 
    \label{eq:vp_Rcjc_i_impl}
\end{equation}
Here, $\mathbf{R}_{b_j b_i}(\mathbf{b}_g)$ represents the bias-corrected relative body rotation integrated from the IMU measurements.

Since VPs correspond to points at infinity, the residual $\mathbf{r}^{\mathrm{vp}}_{ij}$ is inherently translation-invariant, providing an aggregated directional constraint that improves gyroscope bias conditioning. This is particularly effective when point-based cues suffer from low parallax or texture degradation.
Furthermore, regulated by $\mathbb{I}^{\mathrm{vp}}_{ij}$, 
the VP residual is selectively activated only for consistent VP tracks; otherwise, the objective gracefully reduces to the standard point-only formulation.

\subsection{Step 2: Translation-based Linear Alignment with Line-Derived Residuals}
\label{subsec:step2}

Given the gyroscope bias and the refined rotations from Step~1, we perform global linear visual-inertial alignment to estimate the following state vector:
\begin{equation}
    \mathbf{x}=
    \big[
    \ s,\ (\mathbf{g}^w)^\top, (\mathbf{v}_0^w)^\top, \dots, (\mathbf{v}_{N-1}^w)^\top
    \big]^\top,
\label{eq:step2_x_def}
\end{equation}
where $s$, $\mathbf{g}^w$, and $\mathbf{v}_i^w$ denote the metric scale, gravity vector, and the $i$-th keyframe velocity, respectively.

To estimate the state vector $\mathbf{x}$, we formulate the linear alignment as a joint weighted least-squares problem:
\begin{equation}
\begin{aligned}
    \min_{\mathbf{x}}\;&
    \sum_{(i,j)\in\mathcal{E}}
    \Bigg(
    \|\mathbf{r}^{p}_{ij}(\mathbf{x})\|^2_{\mathbf{\Sigma}_{p}} \\
    &\quad +
    \sum_{\ell\in\mathcal{S}_{ij}^{\mathrm{e}}} w^{\mathrm{e}}_{ij,\ell}\, \big(r^{\mathrm{lep}}_{ij,\ell}(\mathbf{x})\big)^2
    +
    \sum_{\ell\in\mathcal{S}_{ij}^{\mathrm{n}}} w^{\mathrm{n}}_{ij,\ell}\, \big(r^{\mathrm{ln}}_{ij,\ell}(\mathbf{x})\big)^2
    \Bigg),
\end{aligned}
\label{eq:step2_joint_objective}
\end{equation}
where $\mathbf{r}^{p}_{ij}$ is the point-based position alignment residual adopted from~\cite{he2023rotation} with covariance $\mathbf{\Sigma}_{p}$. The sets $\mathcal{S}_{ij}^{\mathrm{e}}$ and $\mathcal{S}_{ij}^{\mathrm{n}}$ denote the subsets of visible line tracks used to construct the line epipolar and line-normal residuals, respectively. While $\mathbf{r}^{p}_{ij}$ provides a standard baseline, it often becomes ill-conditioned in the previously discussed degenerate scenarios, where point-based motion estimates yield ambiguous up-to-scale translations $\tilde{\mathbf{p}}_{w c_i}^w$.

To address these limitations, we augment the global alignment with two structureless line-derived residuals that do not require explicit 3D line parameterization: a line epipolar residual $r^{\mathrm{lep}}_{ij,\ell}$ that constrains relative translation through cross-frame line coplanarity, and a line-normal projection residual $r^{\mathrm{ln}}_{ij,\ell}$ that stabilizes the linear system by projecting the baseline position residual $\mathbf{r}^{p}_{ij}(\mathbf{x})$ onto reliable structural directions. 
The formulations of these residuals are detailed in the following subsections.

\subsubsection{Interpretation Plane Normal Estimation from Tracked 2D Lines}
\label{subsubsec:step2_ipn}

To maintain a structureless formulation, each 2D line observation is represented compactly without reconstructing explicit 3D coordinates. Specifically, we parameterize a 2D image line by its interpretation plane. For a tracked line segment $\ell$ in keyframe $i$, its image endpoints are back-projected through the calibrated camera model into projective rays, denoted as $\tilde{\mathbf{s}}_{i,\ell}^{c_i}$ and $\tilde{\mathbf{e}}_{i,\ell}^{c_i} \in \mathbb{R}^3$. The interpretation plane normal in the camera frame is then computed as:
\begin{equation}
    \mathbf{n}_{i,\ell}^{c_i}
    =
    \frac{
    \tilde{\mathbf{s}}_{i,\ell}^{c_i}\times \tilde{\mathbf{e}}_{i,\ell}^{c_i}
    }{
    \left\|
    \tilde{\mathbf{s}}_{i,\ell}^{c_i}\times \tilde{\mathbf{e}}_{i,\ell}^{c_i}
    \right\|
    }
    \in\mathbb{S}^2.
\label{eq:step2_ipn_normal}
\end{equation}
This geometric representation serves as the foundation for formulating the proposed line-derived residuals.

\begin{figure}[t]
    \centering
    \includegraphics[width=0.85\columnwidth]{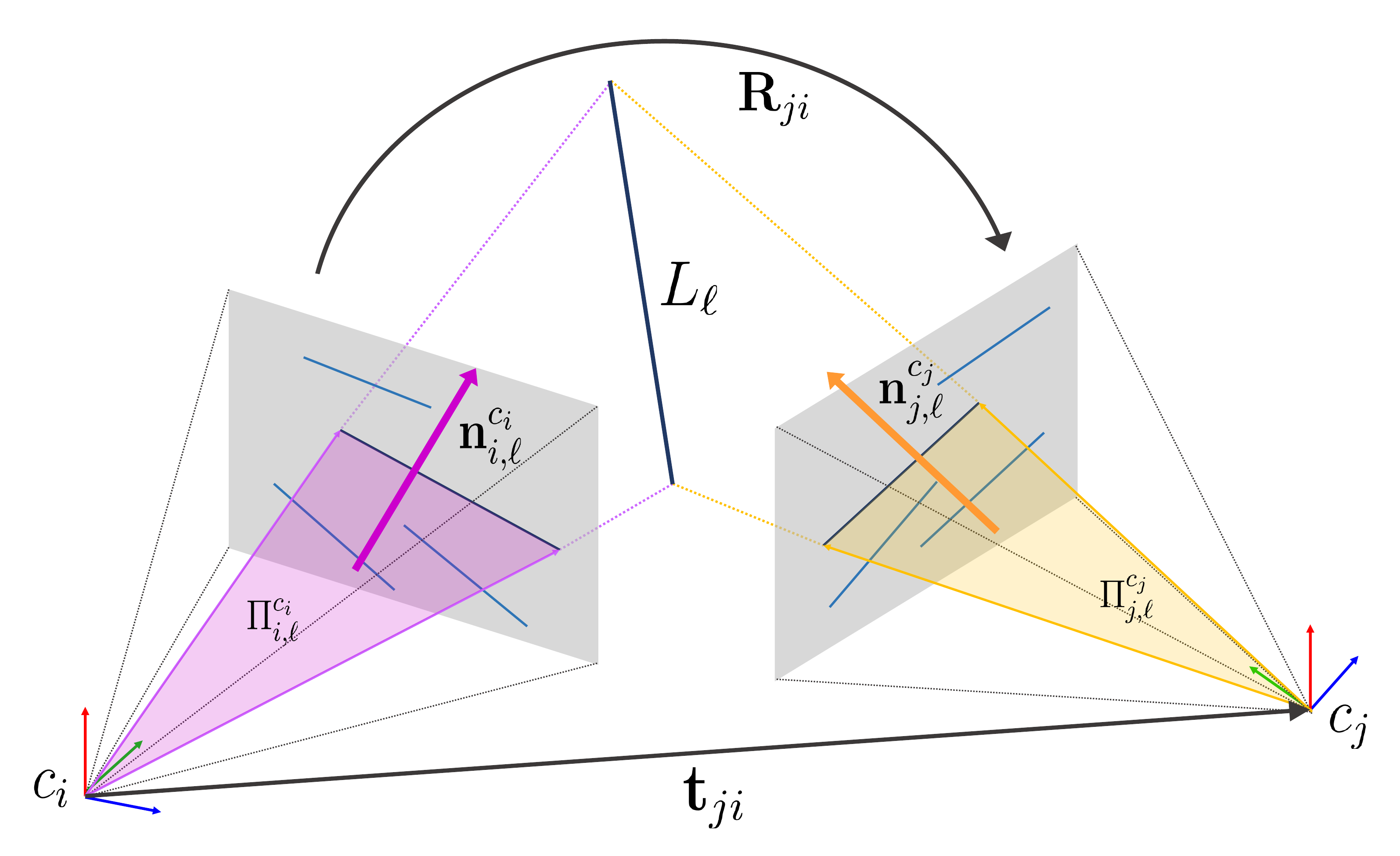}
    \caption{
        Line epipolar geometry between two camera views observing a common 3D line $L_\ell$. The two interpretation planes intersect at the underlying 3D line, and their structural coplanarity with the camera baseline constrains the plane normals as in~(\ref{eq:step2_lineepi_coplanar}).
    }
    \label{fig:method_epipolar}
    \vspace{-0.3cm}
\end{figure}

\subsubsection{Line Epipolar Residual}
\label{subsubsec:step2_lineepi}

For a 3D line observed in both keyframes $i$ and $j$, the corresponding image line measurements impose an epipolar constraint on the relative camera poses.

As shown in Fig.~\ref{fig:method_epipolar}, each line observation, 
along with its camera center, defines an interpretation plane, denoted by $\Pi_{i,\ell}^{c_i}$ and $\Pi_{j,\ell}^{c_j}$. 
For a correct line correspondence, the two planes intersect at the same underlying 3D line. This geometric constraint is expressed as follows:
\begin{equation}
    (\mathbf{n}_{i,\ell}^{c_i})^\top [\mathbf{t}^{c_i}_{ij}]_\times \mathbf{R}_{c_i c_j}\mathbf{n}_{j,\ell}^{c_j} = 0,
\label{eq:step2_lineepi_coplanar}
\end{equation}
where $\mathbf{R}_{c_i c_j}$ and $\mathbf{t}^{c_i}_{ij}$ denote the relative camera rotation and translation from $c_i$ to $c_j$, respectively. Notably, this formulation depends only on 2D line observations and relative motion, bypassing the need for explicit 3D line triangulation.

To incorporate the epipolar constraint into the joint optimization, we express the 
relative translation $\mathbf{t}^{c_i}_{ij}$ as a function of the state vector $\mathbf{x}$. Specifically, each camera position is recovered from its up-to-scale estimate $\tilde{\mathbf{p}}_{w c_i}^w$ using the metric scale $s \in \mathbf{x}$ as $\mathbf{p}_{w c_i}^w = s \tilde{\mathbf{p}}_{w c_i}^w$. The relative translation between the two camera centers, expressed in the $c_i$ frame, is then formulated as:
\begin{equation}
    \mathbf{t}^{c_i}_{ij}(\mathbf{x})
    =
    \mathbf{R}_{c_i w}\big(\mathbf{p}_{w c_j}^w(\mathbf{x})-\mathbf{p}_{w c_i}^w(\mathbf{x})\big). 
\label{eq:step2_tij_from_x}
\end{equation}

By substituting \eqref{eq:step2_tij_from_x} into \eqref{eq:step2_lineepi_coplanar}, we define the line epipolar residual, which quantifies the discrepancy from the ideal coplanarity condition:
\begin{equation}
    r^{\mathrm{lep}}_{ij,\ell}
    \triangleq
    (\mathbf{n}_{i,\ell}^{c_i})^\top
    \big[\mathbf{t}^{c_i}_{ij}(\mathbf{x})\big]_\times
    \mathbf{R}_{c_i c_j}
    \mathbf{n}_{j,\ell}^{c_j}.
\label{eq:step2_lep_residual_direct}
\end{equation}
The proposed line epipolar residual imposes a direct constraint on the scale-dependent relative translation through cross-frame line correspondences. Consequently, it complements point-based constraints and enhances translation observability, particularly in scenarios where point triangulation suffers from insufficient parallax.

\subsubsection{Line-Normal Projection Residual}
\label{subsubsec:step2_linenormal}

To further enhance the linear alignment, we project the point-based position residual $\mathbf{r}^{p}_{ij}(\mathbf{x})$ onto the interpretation plane normals associated with the tracked lines:
\begin{equation}
    {r}^{\mathrm{ln}}_{ij,\ell}(\mathbf{x})
    =
    (\mathbf{n}_{i,\ell}^{b_i})^\top\,\mathbf{r}^{p}_{ij}(\mathbf{x}),
\label{eq:step2_ln_residual}
\end{equation}
where $\mathbf{n}_{i,\ell}^{b_i}$ is the interpretation plane normal expressed in the $i$-th body frame, computed as $\mathbf{n}_{i,\ell}^{b_i} = \mathbf{R}_{b_i c_i} \mathbf{n}_{i,\ell}^{c_i}$. This residual leverages the existing position residual by extracting its component along a geometrically reliable structural direction, thereby enhancing the system's robustness without introducing additional states.

Geometrically, this projection introduces an anisotropic weighting into the alignment objective. In man-made environments, persistent line tracks typically correspond to salient structural boundaries. While point-based translation cues often suffer from isotropic noise under low parallax, the interpretation plane normal of a tracked line provides a robust 1D constraint that suppresses directional ambiguity. By penalizing position errors along these definitive normals, the projection aligns the trajectory with the underlying spatial structure, enhancing the conditioning and numerical stability of the linear solver.

\subsubsection{Line Selection and Adaptive Weighting}
\label{subsubsec:selection_weighting}

Although line-derived residuals provide useful structural cues, the indiscriminate use of all detected lines may introduce noise and computational overhead, potentially compromising both robustness and real-time performance. We therefore select a subset of persistent line tracks based on quality metrics and assign adaptive weights to balance their influence in the optimization. This integration strategy incorporates reliable geometric cues while suppressing noisy or unstable line observations.

For each residual type $t \in \{\mathrm{e, n}\}$, the weight in 
\eqref{eq:step2_joint_objective} is defined as:
\begin{equation}
    w^{t}_{ij,\ell} \propto 0.5 + 0.5 \max(0, \min(q^{t}_{ij,\ell}, 1)),
\label{eq:adaptive_weight}
\end{equation}
where $q^{t}_{ij,\ell} \in [0, 1]$ denotes a normalized residual-type-specific quality score. Specifically, $q^{\mathrm{e}}_{ij,\ell}$ quantifies tracking reliability through sliding-window track persistence, inter-frame matching quality, and valid match ratio, while $q^{\mathrm{n}}_{ij,\ell}$ measures geometric saliency based on normalized line length and detection confidence. This bounded weighting smoothly downweights unreliable features, preserving the numerical stability of the optimization.

\section{Experimental Results}
\label{sec:results}

\begin{figure}[t]
    \centering
    \includegraphics[width=0.95\columnwidth]{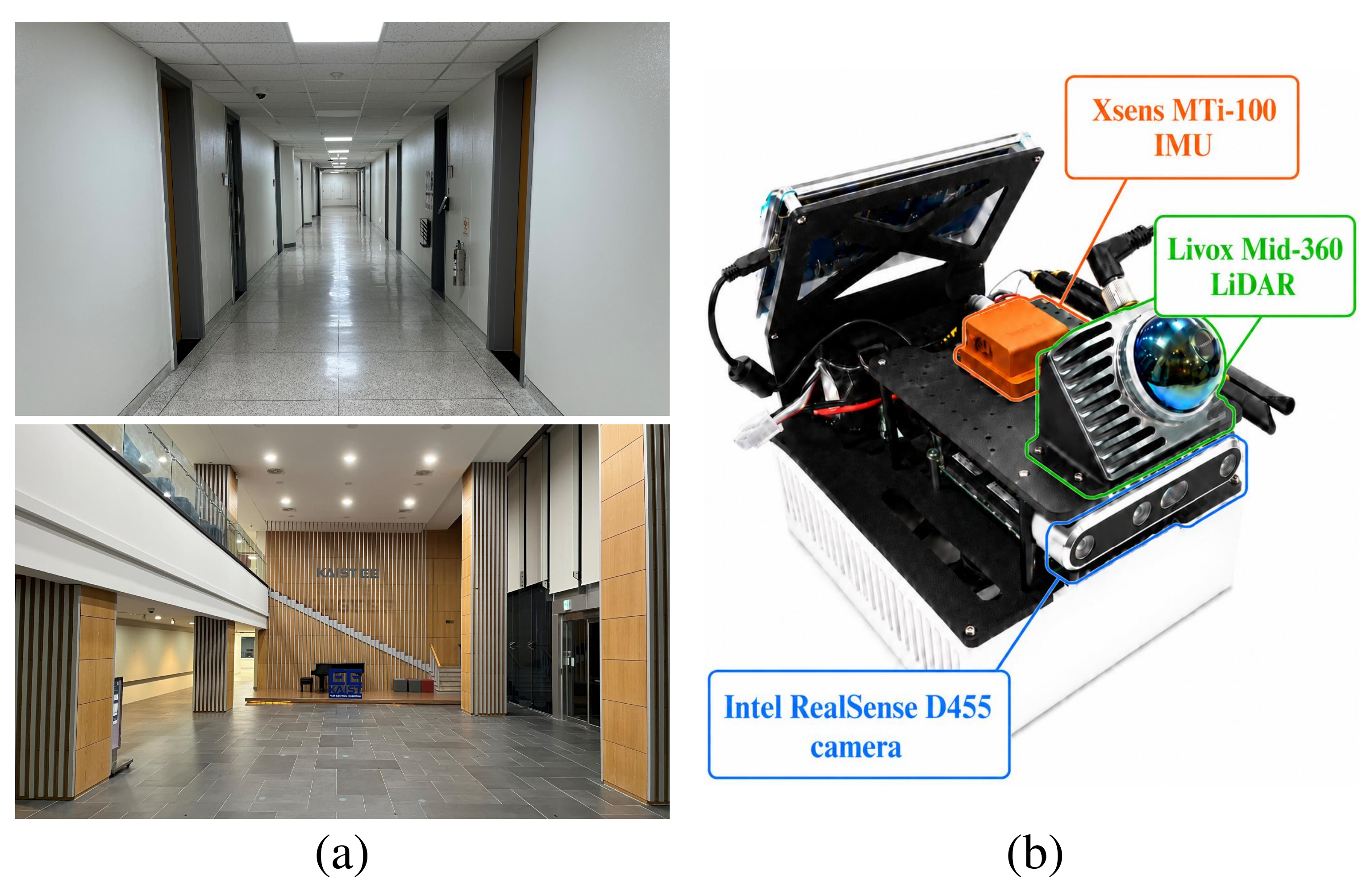}
    \caption{
        Custom degenerate dataset acquisition setup.
        (a) Representative indoor environments in the KAIST E3-2 building, including corridors and a lobby. The collected sequences feature textureless walls, repetitive patterns, low parallax, and translation-dominant motions.
        (b) Sensor module used for data collection.
    }
    \label{fig:custom_data_setup}
    \vspace{-0.3cm}
\end{figure}

\begin{table*}[t!]
    \vspace*{0.05in}
    \setlength{\heavyrulewidth}{1.2pt} 
    \setlength{\lightrulewidth}{0.5pt}
    \centering
    \captionsetup{font=footnotesize}
    \caption{
        Quantitative comparison of exhaustive initialization performance on the EuRoC dataset. \textbf{Bold} and \underline{underlined} values denote the best and second-best results, respectively.
    }
    \label{tab:comparison_euroc}

    \setlength{\tabcolsep}{2.6pt}
    \renewcommand{\arraystretch}{1.05}

    \resizebox{\textwidth}{!}{
    \begin{tabular}{@{}p{32mm}lccccccccccc @{\hspace{4pt}} >{\columncolor{gray!12}}c @{\hspace{4pt}}@{}}
        \toprule
        Metric & Method &
        MH\_01 & MH\_02 & MH\_03 & MH\_04 & MH\_05 &
        V1\_01 & V1\_02 & V1\_03 &
        V2\_01 & V2\_02 & V2\_03 &
        {\hspace{2pt}Avg.\hspace{2pt}} \\
        \midrule

        \multirow{4}{=}{Scale RMSE $\downarrow$} &
        VINS-Mono~\cite{qin2018vins} &
        0.151 & 0.153 & 0.187 & 0.182 & 0.225 &
        0.222 & 0.182 & 0.260 & 0.170 & 0.161 & 0.255 & 0.195 \\
        &
        OpenVINS~\cite{geneva2020openvins} &
        0.137 & \underline{0.121} & \textbf{0.111} & \underline{0.151} & \textbf{0.161} &
        \underline{0.125} & 0.101 & \textbf{0.151} & 0.149 & 0.117 & \textbf{0.118} & \underline{0.131} \\
        &
        DRT-l~\cite{he2023rotation} &
        \underline{0.120} & 0.124 & 0.119 & 0.157 & 0.178 &
        0.127 & \underline{0.091} & \underline{0.187} & \underline{0.137} & \underline{0.112} & \underline{0.142} & 0.136 \\
        &
        Ours &
        \textbf{0.092} & \textbf{0.100} & \underline{0.116} & \textbf{0.144} & \textbf{0.161} &
        \textbf{0.100} & \textbf{0.085} & 0.189 & \textbf{0.104} & \textbf{0.083} & 0.149 & \textbf{0.120} \\
        \midrule
        

        \multirow{4}{=}{Gravity dir. RMSE (\degree) $\downarrow$} &
        VINS-Mono~\cite{qin2018vins} &
        1.409 & 1.130 & 2.261 & 1.940 & 1.629 &
        3.417 & 1.007 & \textbf{1.378} & 1.363 & 1.615 & 1.505 & 1.696 \\
        &
        OpenVINS~\cite{geneva2020openvins} &
        1.931 & 1.558 & 2.239 & 1.758 & 2.326 &
        3.332 & 3.255 & 3.946 & 1.835 & 3.251 & 4.760 & 2.745 \\
        &
        DRT-l~\cite{he2023rotation} &
        \underline{0.948} & \underline{0.968} & \underline{0.917} & \underline{1.062} & \underline{0.943} &
        \underline{3.250} & \textbf{0.914} & 2.547 & \textbf{1.096} & \underline{1.149} & \underline{1.218} & \underline{1.365} \\
        &
        Ours &
        \textbf{0.944} & \textbf{0.933} & \textbf{0.912} & \textbf{0.993} & \textbf{0.907} &
        \textbf{3.224} & \underline{0.937} & \underline{2.158} & \textbf{1.096} & \textbf{1.142} & \textbf{1.176} & \textbf{1.311} \\
        \midrule
        
        \multirow{4}{=}{Velocity RMSE (m/s) $\downarrow$} &
        VINS-Mono~\cite{qin2018vins} &
        0.077 & 0.084 & 0.184 & \textbf{0.148} & 0.200 &
        0.087 & 0.160 & \textbf{0.165} & 0.057 & 0.091 & 0.138 & 0.126 \\
        &
        OpenVINS~\cite{geneva2020openvins} &
        0.153 & 0.111 & 0.250 & 0.186 & 0.200 &
        0.109 & 0.338 & \underline{0.184} & 0.150 & 0.280 & 0.148 & 0.192 \\
        &
        DRT-l~\cite{he2023rotation} &
        \underline{0.060} & \underline{0.066} & \underline{0.131} & 0.192 & \underline{0.163} &
        \underline{0.058} & \underline{0.093} & 0.229 & \underline{0.052} & \underline{0.076} & \underline{0.127} & \underline{0.113} \\
        &
        Ours &
        \textbf{0.054} & \textbf{0.056} & \textbf{0.126} & \underline{0.162} & \textbf{0.140} &
        \textbf{0.042} & \textbf{0.084} & 0.209 & \textbf{0.039} & \textbf{0.068} & \textbf{0.116} & \textbf{0.100} \\
        \midrule
        
        \multirow{4}{=}{Pose RMSE (m) $\downarrow$} &
        VINS-Mono~\cite{qin2018vins} &
        0.074 & 0.087 & 0.180 & \textbf{0.151} & 0.205 &
        0.078 & 0.171 & 0.172 & \underline{0.054} & \underline{0.083} & \underline{0.116} & \underline{0.125} \\
        &
        OpenVINS~\cite{geneva2020openvins} &
        0.103 & 0.087 & \underline{0.163} & \underline{0.179} & \underline{0.188} &
        0.073 & 0.174 & \textbf{0.117} & 0.117 & 0.147 & \textbf{0.109} & 0.132 \\
        &
        DRT-l~\cite{he2023rotation} &
        \underline{0.071} & \underline{0.081} & \underline{0.163} & 0.260 & 0.213 &
        \underline{0.067} & \underline{0.102} & 0.157 & 0.056 & 0.084 & 0.134 & 0.126 \\
        &
        Ours &
        \textbf{0.063} & \textbf{0.069} & \textbf{0.161} & 0.214 & \textbf{0.185} &
        \textbf{0.048} & \textbf{0.084} & \underline{0.155} & \textbf{0.041} & \textbf{0.067} & 0.123 & \textbf{0.110} \\
        \bottomrule
    \end{tabular}%
    }
    \vspace{-0.1cm}
\end{table*}

\begin{figure*}[t]
    \centering
    \includegraphics[width=\textwidth]{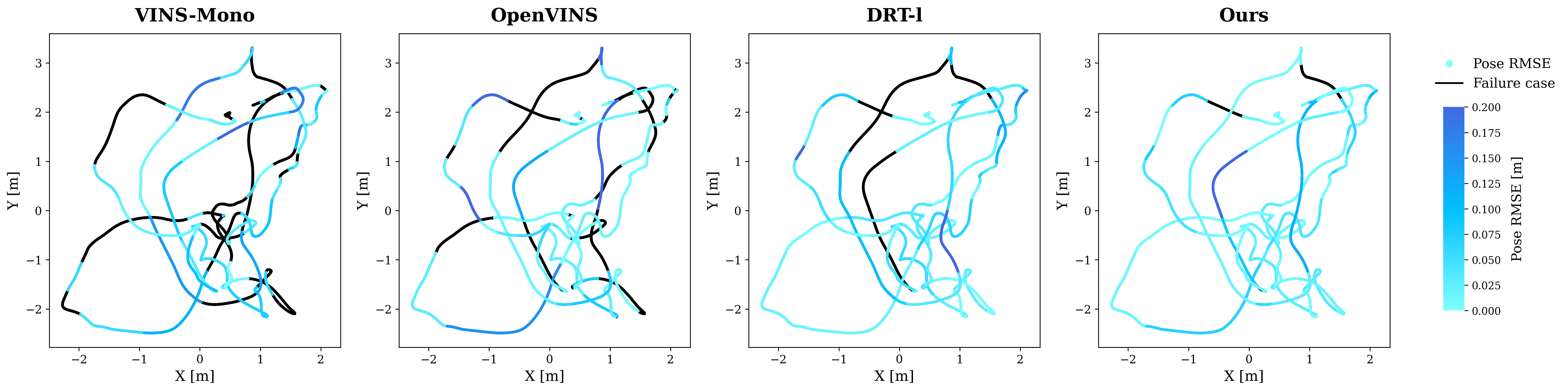}
    \caption{
        Qualitative comparison of initialization performance on the V1\_01 sequence of the EuRoC dataset. Trajectory segments are colored by the pose RMSE (m) computed for each initialization window, where lighter colors indicate lower errors.
        Black segments denote failed initializations. Overall, SLIM-init achieves the fewest failures and the lowest pose errors among the evaluated methods. 
    }
    \label{fig:pose_error}
    \vspace{-0.3cm}
\end{figure*}

\subsection{Baselines and Datasets}

We compare SLIM-init against three representative baselines discussed in Section~\ref{sec:related}. 
For structure-based approaches, we evaluate the loosely-coupled initialization of VINS-Mono~\cite{qin2018vins} and the tightly-coupled closed-form method~\cite{dong2012estimator} implemented in OpenVINS~\cite{geneva2020openvins}.
For the structureless baseline, we consider the DRT method~\cite{he2023rotation} with the loosely-coupled (DRT-l) and tightly-coupled (DRT-t) variants. We adopt DRT-l, as it is shown to achieve better accuracy and computational efficiency than DRT-t~\cite{he2023rotation}. For compactness, SLIM-init is denoted as Ours in all tables and figures.

We evaluate all methods on the EuRoC MAV dataset~\cite{burri2016euroc}, a standard public benchmark for visual-inertial odometry, using its left camera stream for our monocular setup. To assess robustness beyond standard benchmark conditions, we additionally evaluate all methods on custom degenerate-motion sequences. As shown in Fig.~\ref{fig:custom_data_setup}, the custom dataset was collected in the lobby and corridors of the KAIST E3-2 building using an Intel RealSense D455 camera and an Xsens MTi-100 IMU. Ground-truth trajectories for the custom dataset were generated using FAST-LIO2~\cite{xu2022fastlio2}.

\subsection{Evaluation Metrics}
\label{subsec:metrics}

We evaluate initialization accuracy based on errors in scale, gravity direction, velocity, and pose.
To compute the scale errors, the estimated trajectory is first aligned to the ground truth via a $\mathrm{Sim}(3)$ transformation~\cite{Umeyama1991least}.
For all metrics, we report the root mean square error (RMSE) over successful initializations only, where success is defined by a bounded scale estimate, $|s-1|<0.5$.
In addition, to validate the effectiveness of the proposed VP residual in our ablation study, we evaluate the gyroscope bias error.
Following~\cite{he2023rotation}, we compute the percent relative error in gyroscope bias magnitude, $e_{\text{gyro}}(\%) = 100 \cdot
\{{\left|\|\mathbf{b}_{g}\|-\|\bar{\mathbf{b}}_{g}\|\right|} \ / \  {\|\bar{\mathbf{b}}_{g}\|}\}$,
where $\bar{\mathbf{b}}_{g}$ denotes the ground truth gyroscope bias provided by the dataset.
Finally, we report the initialization success rate as a measure of robustness, and the average initialization solve time to quantify computational efficiency.

\subsection{Experimental Setup}

To rigorously analyze initialization performance, we adopt an exhaustive testing protocol similar to prior work~\cite{he2023rotation}.
Specifically, we launch an initialization trial every $10$ raw image frames by shifting the trial start time with a fixed stride of $10$ frames.
For each trial, we form a sliding window by selecting $10$ keyframes from the corresponding segment.
This yields densely overlapping evaluation windows throughout each sequence, reflecting practical deployment where re-initialization may be triggered at arbitrary times due to unexpected hardware/software faults or degraded state estimation quality. 
All experiments are conducted on an Intel Core i9-12900K CPU and 32GB of RAM.


\subsection{Initialization Accuracy Analysis}

Table~\ref{tab:comparison_euroc} summarizes the initialization accuracy on the EuRoC dataset.
Overall, SLIM-init achieves the best average performance across all evaluated metrics. 
Also, SLIM-init ranks first or second in most individual sequences, highlighting its reliability under diverse conditions. 
In particular, compared with the baseline DRT-l~\cite{he2023rotation}, SLIM-init reduces the average RMSE of scale, gravity direction, and velocity estimation by 11.8\%, 4.0\%, and 11.5\%, respectively. 
Pose estimation also benefits from the improved alignment stability, resulting in the lowest average pose RMSE of 0.110, which is 12.7\% lower than DRT-l and the lowest among all evaluated methods.
This indicates that reinforcing both rotation and translation observability through line-derived cues improves initialization conditioning across diverse benchmark sequences.
The effectiveness of SLIM-init is further shown in Fig.~\ref{fig:pose_error}, where it exhibits fewer failures and lower pose errors than the baselines.

\subsection{Robustness Analysis Under Degenerate Conditions}

Beyond average accuracy, Table~\ref{tab:success_rate} reports the initialization success rate on the EuRoC dataset. While structure-based baselines often fail on sliding windows corresponding to trajectory segments with challenging conditions, SLIM-init consistently achieves the highest success rates across sequences.

To further validate robustness under more severe degeneracies, we evaluate all methods on the custom degenerate-motion sequences, as summarized in Table~\ref{tab:comparison_custom}.
On these sequences, the structure-based methods (VINS-Mono and OpenVINS) fail to initialize.
While the structureless point-based method (DRT-l) remains operational, SLIM-init outperforms this baseline across all test sequences, indicating improved robustness under severe degeneracies.
These results suggest that the proposed line-derived structural constraints facilitate reliable metric initialization even under degenerate scenarios.

\begin{table}[t]
    \vspace*{0.05in}
    \centering
    \captionsetup{font=footnotesize}
    \caption{
        Initialization success rates on the EuRoC dataset.
        Best results are marked in \textbf{bold}.
    }
    \label{tab:success_rate}
    
    \setlength{\tabcolsep}{2pt}
    \resizebox{\columnwidth}{!}{
    \begin{tabular}{@{}lcccc@{}}
        \toprule
        \multirow{2}{*}{Sequence} &
        \multicolumn{4}{c}{Success rate (\%) $\uparrow$} \\
        \cmidrule(lr){2-5}
        & VINS-Mono~\cite{qin2018vins} & OpenVINS~\cite{geneva2020openvins} & DRT-l~\cite{he2023rotation} & {\hspace{4pt}} Ours {\hspace{2pt}} \\
        \midrule

        Machine Hall & 63.84 & 60.22 & 91.77 & \textbf{94.01} \\
        Vicon Room 1 & 22.97 & 62.90 & 72.08 & \textbf{73.14} \\
        Vicon Room 2 & 28.78 & 59.35 & 71.51 & \textbf{72.11} \\
        \bottomrule
    \end{tabular}
    }
    \vspace{-0.1cm}
\end{table}

\begingroup
\makeatletter
\setlength{\heavyrulewidth}{1.2pt} 
\setlength{\lightrulewidth}{0.5pt} 
\makeatother
\begin{table}[t]
    \centering
    \captionsetup{font=footnotesize}
    \caption{
        Quantitative comparison of exhaustive initialization performance on custom degenerate sequences. 
        Best results are marked in \textbf{bold}.
    }
    \label{tab:comparison_custom}

    \setlength{\tabcolsep}{2.6pt}
    \renewcommand{\arraystretch}{1.05}

    \resizebox{\columnwidth}{!}{
    {\scriptsize
    \begin{tabular}{@{}p{16mm}lccc @{~~~} >{\columncolor{gray!12}}c @{~}}
        \toprule
        Metric & Method &
        {~~C\_01~~} & {~~C\_02~~} & {~~C\_03~~} & { Avg. } \\
        \midrule

        \multirow{2}{=}{Scale RMSE $\downarrow$} &
        DRT-l~\cite{he2023rotation} &
        0.223 & 0.132 & \textbf{0.144} & 0.166 \\
        &
        Ours &
        \textbf{0.208} & \textbf{0.101} & \textbf{0.144} & \textbf{0.151} \\
        \midrule

        \multirow{2}{=}{Gravity dir. \\ RMSE (\degree) $\downarrow$} &
        DRT-l~\cite{he2023rotation} &
        5.269 & 2.550 & 4.817 & 4.212 \\
        &
        Ours &
        \textbf{5.154} & \textbf{2.507} & \textbf{4.751} & \textbf{4.137} \\
        \midrule

        \multirow{2}{=}{Velocity \\ RMSE (m/s) $\downarrow$} &
        DRT-l~\cite{he2023rotation} &
        0.346 & 0.176 & 0.239 & 0.254 \\
        &
        Ours &
        \textbf{0.335} & \textbf{0.152} & \textbf{0.233} & \textbf{0.240} \\
        \midrule

        \multirow{2}{=}{Pose \\ RMSE (m) $\downarrow$} &
        DRT-l~\cite{he2023rotation} &
        0.359 & 0.119 & \textbf{0.169} & 0.216 \\
        &
        Ours &
        \textbf{0.355} & \textbf{0.095} & \textbf{0.169} & \textbf{0.206} \\
        \bottomrule
    \end{tabular}%
    }
    }
    \vspace{-0.3cm}
\end{table}
\endgroup

\subsection{Solve Time and Computational Efficiency}
\label{subsec:solve_time}

\begin{figure}[t]
    \centering
    \includegraphics[width=0.87\columnwidth]{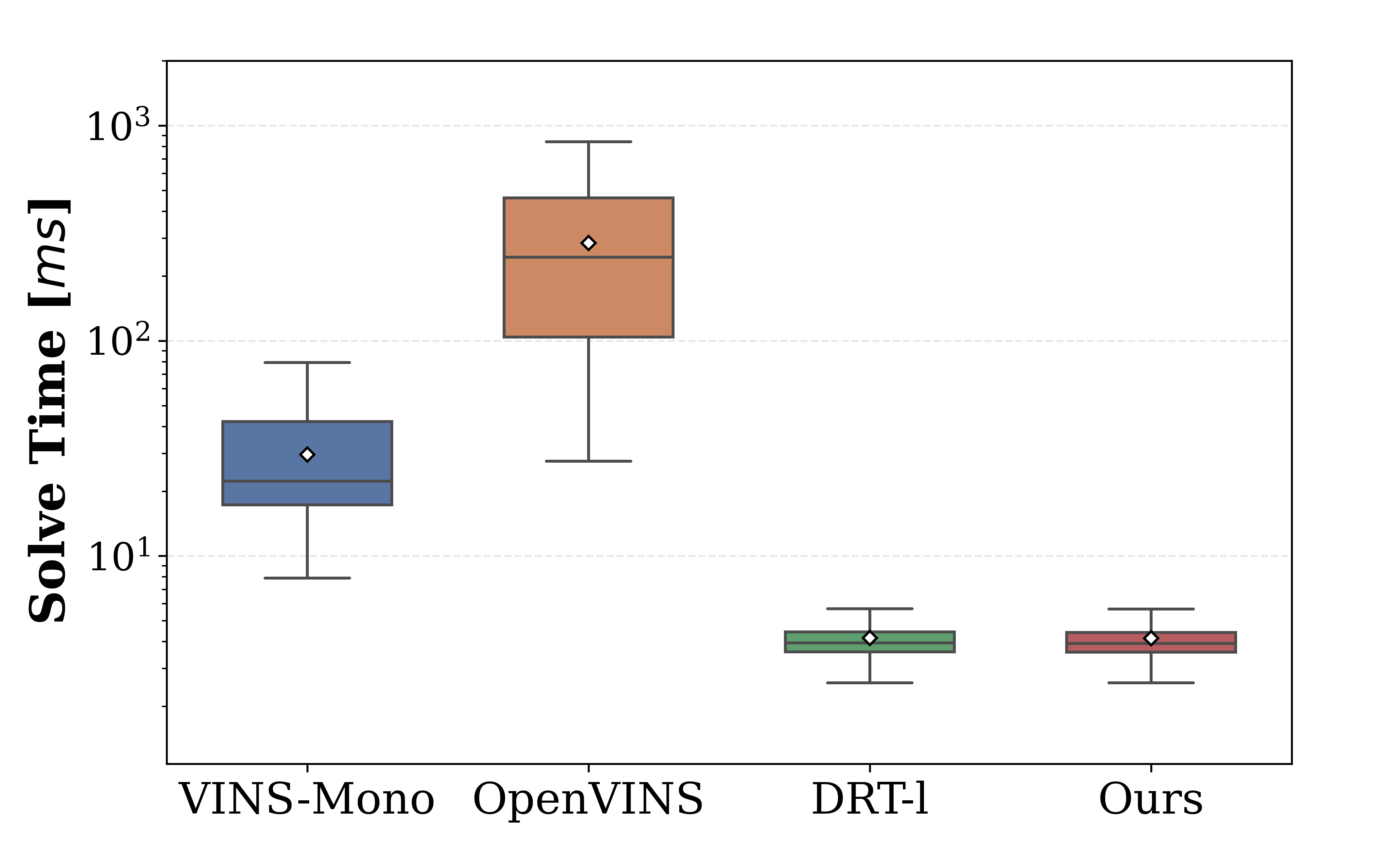}
    \caption{Comparison of initialization solve time on the EuRoC dataset.
    The central line denotes the median, and diamond markers denote the mean solve time.}
    \label{fig:solve_time}
    \vspace{-0.1cm}
\end{figure}

Fig.~\ref{fig:solve_time} reports the initialization solve time on the EuRoC dataset. 
With an average of 4.141~ms, SLIM-init achieves a solve time similar to the 4.150~ms of the structureless point-based baseline, DRT-l~\cite{he2023rotation}, indicating negligible computational overhead from the additional residuals.
Moreover, it is substantially faster than conventional structure-based pipelines such as VINS-Mono~\cite{qin2018vins} and OpenVINS~\cite{geneva2020openvins}, which require 29.615~ms and 284.715~ms, respectively. These results show that the proposed residuals incorporate structural constraints while maintaining a compact state vector for efficient linear alignment.

Although SLIM-init introduces marginal overhead over the point-only baseline~\cite{he2023rotation} due to additional 2D line tracking and VP estimation, this cost remains tightly bounded.
As feature tracking constitutes the primary bottleneck, we limit the 
additional load by using a single dominant VP per frame and a capped number of persistent line tracks.
Consequently, the overall initialization requires only 35.86~ms per image frame on average, preserving real-time performance for fast startup and frequent recovery.

\begingroup
\makeatletter
\setlength{\heavyrulewidth}{1.2pt}
\setlength{\lightrulewidth}{0.5pt}
\makeatother
\begin{table}[t]
    \centering
    \captionsetup{font=footnotesize}
    \caption{Ablation study on custom degenerate sequences, reported as averages across all degenerate sequences. 
    (Sca.: scale RMSE, Grav.: gravity direction RMSE~($^\circ$), Vel.: velocity RMSE~(m/s), Pos.: pose RMSE~(m)). 
    Best results are marked in \textbf{bold}.}
    \label{tab:ablation_custom_avg}

    \setlength{\tabcolsep}{5pt}
    \renewcommand{\arraystretch}{1.05}

    \resizebox{\columnwidth}{!}{%
    \begin{tabular}{@{}lcccc@{}}
        \toprule
        Method &
        Sca. $\downarrow$ &
        Grav. $\downarrow$ &
        Vel. $\downarrow$ &
        Pos. $\downarrow$ \\
        \midrule
        Ours w/o\ both line residuals                & 0.166 & 4.212 & 0.254 & 0.216 \\
        Ours w/o\ line epipolar residual             & 0.162 & 4.142 & 0.245 & 0.215 \\
        Ours w/o\ line normal residual               & 0.158 & 4.196 & 0.245 & 0.212 \\
        \midrule
        Ours      & \textbf{0.151} & \textbf{4.137} & \textbf{0.240} & \textbf{0.206} \\
        \bottomrule
    \end{tabular}%
    }
    \vspace{-0.3cm}
\end{table}
\endgroup

\subsection{Ablation Study}
We conduct an ablation study to validate the individual contributions of the proposed VP and line-derived residuals.

\subsubsection{Effect of the VP residual in Step 1}
On the EuRoC dataset, disabling the VP residual in Step~1 increases the average gyroscope bias error~$e_{\text{gyro}}$ from 1.913\% to 1.941\%. This shows that translation-invariant VP cues effectively complement point-based rotation constraints by providing stable directional measurements.

\subsubsection{Effect of the line residuals in Step 2}
Table~\ref{tab:ablation_custom_avg} summarizes the impact of the line epipolar and line-normal projection residuals in Step~2 on custom degenerate sequences. Removing both residuals results in degradation across all metrics, validating the importance of incorporating line-derived constraints in degenerate settings. 

Removing either residual also leads to performance degradation. Notably, the absence of the line-normal projection residual results in the highest gravity RMSE ($4.196^\circ$), underscoring its pivotal role in stabilizing gravity direction estimation during linear alignment. Furthermore, the line epipolar residual proves more critical for scale recovery, as its removal yields a larger scale error ($0.162$) compared to excluding the line-normal projection term ($0.158$). Overall, the full model achieves superior performance, confirming that these two residuals complement each other to enhance both geometric consistency and metric accuracy.

\section{Conclusion}
\label{sec:conclusion}


We proposed SLIM-init, a structureless monocular visual–inertial initialization method that incorporates line-derived geometric constraints without requiring explicit 3D reconstruction. By leveraging structural cues from vanishing points and 2D line-derived constraints, SLIM-init improves initialization robustness while preserving real-time performance. Experimental results demonstrate improved accuracy and robustness over state-of-the-art approaches. Future work will explore adaptive structural modeling across diverse environments and tighter integration of line-derived constraints into VIO backends to further improve long-term stability.




\bibliographystyle{URL-IEEEtrans}

\bibliography{URL-bib}

\end{document}